\documentclass[a4paper,10pt]{article}
\usepackage[utf8]{inputenc}
\usepackage{url}
\usepackage{xcolor}
\usepackage[a4paper, left=1.4cm, right=1.4cm, top=2cm, bottom=2.5cm]{geometry}
\usepackage[linkcolor=blue, citecolor=blue, urlcolor=blue, colorlinks=true]{hyperref}
\usepackage{multicol}
\usepackage{graphicx}
\usepackage[affil-it]{authblk}
\usepackage{hyperref}
\usepackage{titlesec}
\usepackage{float}
\usepackage{lipsum}  
\usepackage{subfig}
\usepackage{mwe}
\usepackage[english]{babel}

\titleformat*{\section}{\large\bfseries}
\titleformat*{\subsection}{\normalsize\bfseries}
\titleformat*{\subsubsection}{\small\bfseries}

\addto\extrasenglish{%
}

\title{\textbf{AU-NN: ANFIS Unit Neural Network}}
\author{\normalsize Tonatiuh Hernández-del-Toro \thanks{Email: \textbf{tonahdztoro@gmail.com}} }
\author{\normalsize Carlos A. Reyes-García}
\author{\normalsize Luis Villaseñor-Pineda}
\affil{\footnotesize Biosignal Processing and Medical Computing Lab. Instituto Nacional de Astrofísica, Óptica y Electrónica. Mexico.}

\date{}

\begin{document}

\maketitle

\noindent\makebox[\linewidth]{\rule{\textwidth}{0.4pt}} 
\vspace{0cm}\\
\textbf{Abstract}
\vspace{0.2cm}\\
\noindent In this paper is described the ANFIS Unit Neural Network, a deep neural network where each neuron is an independent ANFIS. Two use cases of this network are shown to test the capability of the network. (i) Classification of five imagined words. (ii) Incremental learning in the task of detecting Imagined Word Segments vs. Idle State Segments. In both cases, the proposed network outperforms the conventional methods. Additionally, is described a process of classification where instead of taking the whole instance as one example, each instance is decomposed into a set of smaller instances, and the classification is done by a majority vote over all the predictions of the set. The codes to build the AU-NN used in this paper, are available on the github repository \url{https://github.com/tonahdztoro/AU_NN}.
\vspace{0.2cm}
\noindent\makebox[\linewidth]{\rule{\textwidth}{0.4pt}}

\section{Introduction}
In the real physical world, most of things are not discrete, but lay on a continuous space. This is also the way humans process the reality. However, current machines do not think in a continuous way, but discrete. If we desire to build machines that think more like humans, we must find other paradigms more suitable. Fuzzy systems offer a method of processing information in a continuous way, which is more likely to the physical reality. Several implementations of fuzzy systems have been developed to successfully face complex problems \cite{Kar2014, Das2020}. Nevertheless, there are still datasets which are difficult to deal with, and have not been fully solved yet. Among them, is the imagined speech in the EEG signals \cite{Brigham2010}, they have a chaotic and non linear behavior, this kind of behavior is well processed by fuzzy systems \cite{Lin1996}. 

One of the most popular implementations of fuzzy systems is the ANFIS \cite{Jang1993}, which is a combination of neural networks and fuzzy inference systems. The implementation of ANFIS has proven to be efficient in different areas \cite{Sengur2008, Atsalakis2009, Chakrapani2010, Galavi2012}. However, in its early stages of implementations it required a lot of computational resources to be implemented on large datasets.

In recent years, due to the parallel computing using GPU clusters, it has been possible to solve problems that prior required a lot of computational resources, the most popular use of GPU in machine learning are the deep neural networks \cite{Ekman2021}. The computational power of GPU can be used to solve bigger problems of fuzzy systems as the ANFIS in large datasets or other implementations of these types of systems.

Another problem that is faced in EEG signals and Brain Computer Interfaces implementations is the Incremental Learning (IL), which refers to the situation of continuous model adaptation based on a constantly arriving data-stream \cite{Losing2018, Gepperth2016, Silver2011}. This is because EEG signals are naturally changing over time, and also, sometimes the process of obtaining EEG signals is exhaustive for the subject, leading to smaller datasets that require models that need to be trained with a different paradigm where we have few instances at the beginning and newer instances that are added to the dataset over time. Fuzzy systems can also solve this kind of problems since they have proven to be adaptive to incremental learning \cite{Almaksour2012, Matsumura2017}.

In this paper, the ANFIS Unit Neural Network (AU-NN) is presented. The AU-NN is an artificial neural network that exploits the benefits of fuzzy systems like the ANFIS and Deep learning with tools as GPU parallel computing. This network is tested with two problems that involve EEG signals. (i) The classification of imagined words in EEG, and (ii) The detection of Imagined Word Segments (IWS) vs. Idle State Segments (ISS) \cite{Hernandez-Del-Toro2021} in an incremental learning way. In addition, a process of classification is described, where instead of taking the whole instance and taking it as one example, each instance is decomposed into a set of smaller instances, and the classification is done by a majority vote over all the predictions of the set.

The results for the classification of imagined words showed that the AU-NN outperforms the classification results of previous works on the dataset recorded in \cite{Torres-Garcia2016}. For the IL part, as there is no previous work reported that involves imagined speech using IL, The AU-NN is tested against traditional models as the RF classifier which also outperforms.

The rest of the paper is organized as follows: In \autoref{sec:RW}, a review of the previous works related to each topic is done. \autoref{sec:AU-NN} describes how the AU-NN is built and how it works. In \autoref{sec:IWC} is described the experiments, and discussion of the results obtained with the classification of imagined words using the AU-NN. In \autoref{sec:IL} is described the experiments, and discussion of the results obtained with the IL AU-NN on the task of detecting IWS vs. ISS. Finally, in \autoref{sec:conclusion} a conclusion is given.

\section{Related work}\label{sec:RW}
Since the presentation of the fuzzy systems, several implementations have been made \cite{Jang1993, Jang1997}. The fuzzy systems have been used to successfully solve a variety of problems of different areas \cite{Kar2014, Das2020}: medical datasets \cite{Sengur2008, Ubeyli2009, Kumar2011, Agboizebeta2012}, economics \cite{Atsalakis2009, Giovanis2012, Fang2012}, image processing \cite{Chakrapani2010, Gomathi2010}, forecasting \cite{Galavi2012, Nayak2005, Patil2011}, among many others.

In the study of imagined speech in EEG signals \cite{Brigham2010}, there have been several works that intend to solve the decoding of imagined speech on brain signals \cite{Panachakel2021}. Among these works there are research that aims to classify vowels \cite{Min2016, Nguyen2017, Saha2019, Cooney2020}, words related to directions \cite{Torres-Garcia2016, Garcia-Salinas2018, Moctezuma2017, Alsaleh2017, Cooney2020}, and words to answer yes/no questions \cite{Sereshkeh2017, Balaji2017}. These works have shown promising results. However, none of them has successfully built a functional online BCI. This is due to the complexity of the problem of dealing with imagined speech.

In the IL, although the problem has not been fully solved in a general way \cite{Losing2018, Gepperth2016, Silver2011}, several incremental learning algorithms have been proposed \cite{Liang2006, Cauwenberghs2001, Bordes2005, Losing2015, Polikar2001, Saffari2009}. However, this algorithms only work in a specific task scenario.

\section{The ANFIS Unit Neural Network}\label{sec:AU-NN}
In a vanilla neural network, each neuron in each layer is a non-linear unit that takes the feature vector as input, multiplies the features by some weights, sum the results, and inputs that sum into a non-linear function which can be a tanh, ReLU, or a sigmoid function, in the latter case, the neuron becomes a logistic regression unit. The output of all neurons in one layer becomes the feature vector for the next layer. This process is repeated until we reach the final layer which will decide the class of the instance. A deep neural network consists of this model  but with many layers and many neurons in each layer along with many other properties, attempting with this deepness to extract features in each layer and to successfully generalize over a dataset \cite{Ekman2021}.

Similarly to the construction of a deep neural network, we define the ANFIS Unit Neural Netowrk (AU-NN) as a deep neural network where each neuron, instead of computing the process described above, are independent ANFIS \cite{Jang1993}. Since the ANFIS is complex enough to mimic non-linear functions \cite{Jang1997}, we do not need to introduce extra non-linearities. We can use each independent ANFIS as a neuron, being part of a bigger deep neural network and apply and argmax in the last layer to obtain the class. 

\begin{figure}[ht]
    \centering
    \includegraphics[width=0.7\textwidth]{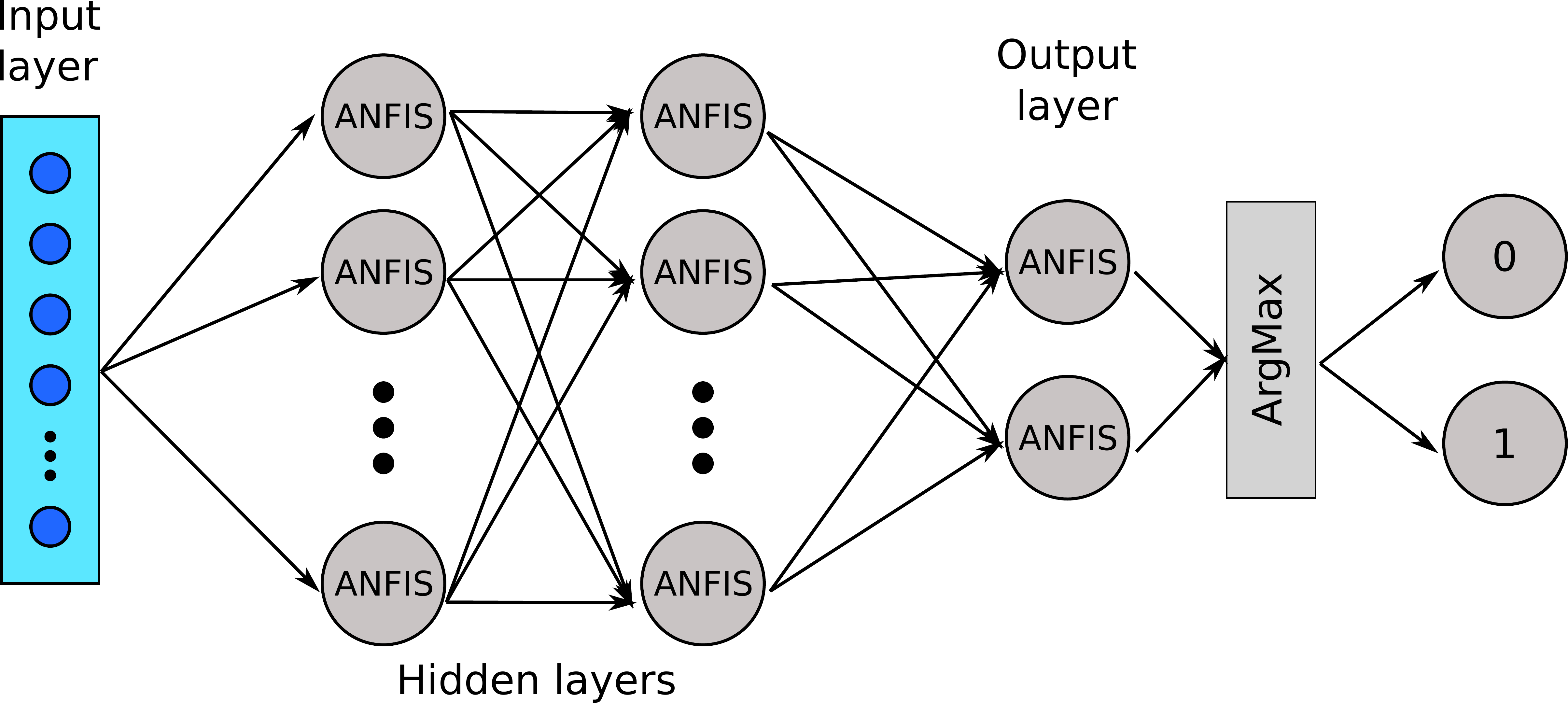}
    \caption{Fully connected AU-NN with two hidden layers for a two class problem, each neuron is an independent ANFIS with predefined number of membership functions, and the full AU-NN can be trained using training algorithms like Adam \cite{Kingma2015}.}
    \label{fig:AU_NN}
\end{figure}

In \autoref{fig:AU_NN}, an example of an AU-NN for a two class problem is shown. Each neuron will be an independent ANFIS. The number and type of membership functions, as well as the number of layers will be defined by the user depending of the problem at hand.

\subsection{How the neural network is trained}
A solely ANFIS by itself is very complex and needs the use of several computational resources. In order to implement a functional AU-NN, we are going to need the use of more computational resources. To solve this problem we make use of the parallel GPU computation, with programming tools like tensorflow, which is widely use for the development of deep neural networks.

As we are using the parallel computing power of tensorflow and GPU's, we no longer need the hybrid learning algorithm used in the traditional ANFIS to speed the training phase. Instead we can treat all the AU-NN as the same optimization problem and make use of other efficient training algorithms like Adam \cite{Kingma2015}.

With this neural network we can face classification problems that require a more complex treatment as the imagined speech on EEG signals.

The codes for the python implementation of the AU-NN are available on the github repository \url{https://github.com/tonahdztoro/AU_NN}.

\section{Imagined word classification using AU-NN}\label{sec:IWC}
In this section is described the experiment performed for imagined word classification on a dataset of 27 subjects using the AU-NN with a majority vote classification scheme.

\subsection{Classification using majority vote}
In a traditional classification problem, an instance is feed to the trained model, and the model attempts to output the correct class.

In this experiment, instead of using the whole instance signal (word), extracting features and aim to classify them, we are going to segment each word into a set of overlapped smaller instances. Each smaller instance will have its own feature vector, and all of those feature vectors from the set corresponding to the word will be classified, the final class is selected as the one that is more predicted among all the segments in the set.

For the training part, similarly, the instances that belong to the training set will be segmented into sets of smaller instances, and then, the model will learn from this bigger training set containing all the instances of each word.

\begin{figure}[ht]
    \centering
    \includegraphics[width=0.9\textwidth]{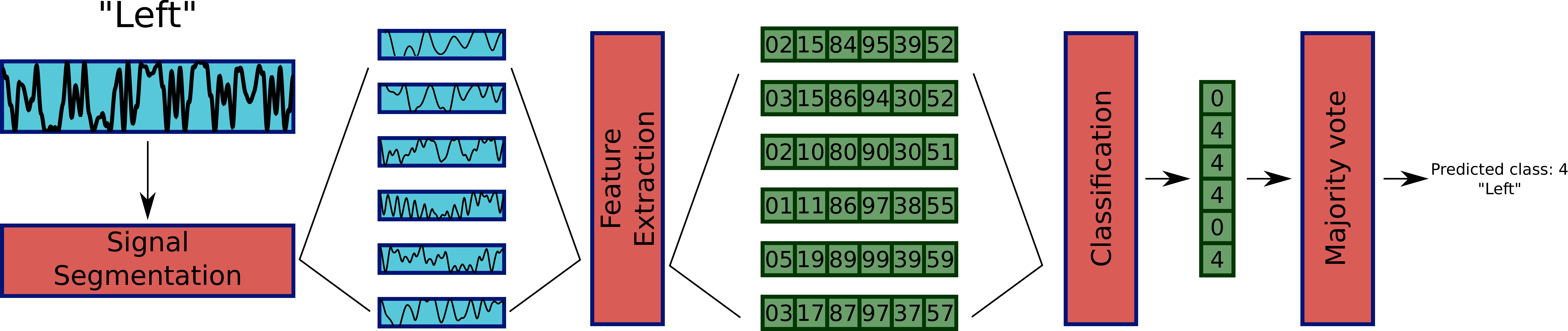}
    \caption{Example of a majority vote classification. Instead of extracting only one feature vector from an instance, the instance is divided into overlapped segments, and each of those segments will become a smaller feature vector, the model learns to classify this segments. For the testing part, from each instance, a set of feature vectors are obtained, and all of them are feed to the model, the class which is predicted the most is the selected for the classification.}
    \label{fig:majority}
\end{figure}

In \autoref{fig:majority} and example of this description is shown.

\subsection{Experiment description}
Using the dataset described on \cite{Torres-Garcia2016}, the AU-NN is tested on a classification problem: five imagined words in Spanish which mean (up, down, left, right, select) with recordings of EEG of 27 subjects. The features used are the Instant Wavelet Energies of each channel, with 4 levels of decomposition, using the bior2.2 wavelet as mother wavelet. For each signal (word), a window of 0.5 seconds is used to extract the set of smaller instances used for the majority vote classification. The window is moved each time 0.1 seconds and a new instance is obtained, this process is repeated for all the words.

A subject-independent classifier is trained for each subject, and 5-fold validation scheme was performed, 80\% of the words are used for training while the rest 20\% are used for the testing set.

Several architectures of the network were tried, and the architecture selected for its best performance is an AU-NN of one layer with number of neurons equal to the number of classes (5). The number of membership functions is set equal to the number of features of the dataset (70). The membership function used is the triangular function. For the optimization parameters, the Adam optimizer was used, with a learning rate schedule chosen as [0.03, 0.01, 0.003], and 500 epochs per learning rate were performed.

\subsection{Results}
\begin{figure}[ht]
    \centering
    \includegraphics[width=0.8\textwidth]{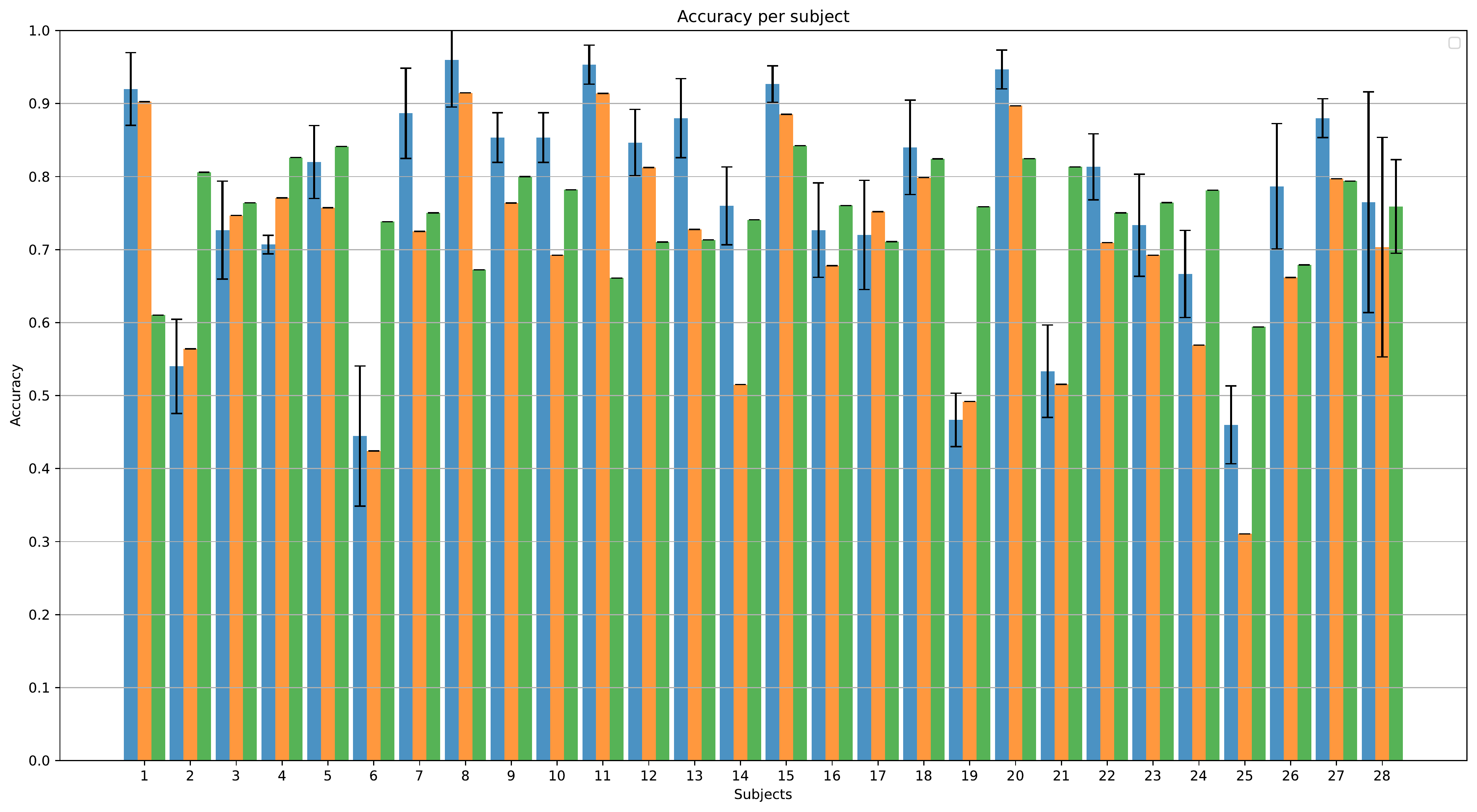}
    \caption{Accuracies obtained using the AU-NN with the imagined words dataset, the blue bars represent the accuracy obtained for each subject using the proposed AU-NN, the orange bars are the accuracies obtained in \cite{Torres-Garcia2016}, and the green bars represent the accuracies obtained in \cite{Garcia-Salinas2018}. The blue bars contain a standard deviation line, which represents the variation in folds, the standard deviation in the mean bar (28) represents the standard deviation among all subjects.}
    \label{fig:imagined_words}
\end{figure}

The classification results are shown in \autoref{fig:imagined_words}, and the obtained accuracies are compared with the results reported in \cite{Torres-Garcia2016}, and \cite{Garcia-Salinas2018} for the same dataset.

\subsection{Discussion}
For 15 subjects, the AU-NN outperforms the previous reported methods, in 7 subjects, the AU-NN performs similarly, and in 5 subjects the AU-NN performs badly. Although the standard deviation for the mean of all the subjects is big for the AU-NN, the individual standard deviation for each subject is smaller which behaves as a more robust classifier. 

It is also important to notice that in the subjects that the AU-NN performs badly, the accuracy is similar or better as the ones reported in \cite{Torres-Garcia2016}. This can be explained because in the latter work, they used similar feature extraction methods (Wavelets), although they used a Random Forest classifier.

On the other hand, the results in which the AU-NN performs drastically badly against those reported in \cite{Garcia-Salinas2018} can be due to the feature extraction used in that work, they used a tensor decomposition, and an SVM classifier. Perhaps using this feature extraction method along with the AU-NN, the performance for all subjects could increase.

With this results, we prove the capability of the AU-NN to perform good in a complex classification problem.

\section{AU-NN applied to IWS vs. ISS with an IL approach}\label{sec:IL}
In this section is described the experiment performed for IL using the AU-NN for the detection of IWS vs. ISS.

\subsection{Experiment description}
This is a two class classification problem with IL. To realize this task, the method of signal segmentation used in \cite{Hernandez-Del-Toro2021} was applied to obtain the exact same IWS and ISS instances. From the three datasets used in that work, the first dataset, the one recorded in \cite{Torres-Garcia2016} was selected, this is because it is the only one from the three datasets that was recorded in sessions, this is a characteristic that we need in order to train and test the model in an IL approach. From all the feature sets used in \cite{Hernandez-Del-Toro2021} the one based on Instant Wavelet Energies (IWE) was selected, this is because for the given dataset this feature set was the one that reported the higher results.

A detailed description of the experiment is given below:
\begin{enumerate}
    \item There are 27 subjects with 5 sessions each one. Each session contains 20 trials.  This leads to 100 trials per subject.
    \item From each session, 50\% of trials are randomly selected to build a train session set, and the other 50\% are kept for test set.
    \item The trials extracted for the train session sets remain in their respective sessions. The trials extracted from the test set are mixed over all sessions.
    \item For the train sets, from each trial in each session, segments are extracted with a moving window of size 0.5 seconds and an overlap of 0.1 seconds, and features are extracted as in \cite{Hernandez-Del-Toro2021}. With this, each trial is turned into a set of instance vectors of IWS's and ISS's. Here is important to mention that in this experiment, the classification using majority vote was not applied, and each instance was taken independently as an example.
    \item With this, the test set is composed of the instances extracted from 50 trials (10 trials per session). And train session sets consist of the instances extracted from the 5 sets of trials (10 trials per session).
    \item With this procedure, each instance has 70 features, and the number of training instances per sessions will vary in every session, although the number of instances in the test set remains the same.
    \item With the train sets of each session, the AU-NN is trained one session at a time. After training each session, the model is tested with the whole test set. As each session represents a different word, and the model does not classifies words, but learns to detect between IWS vs ISS, the expected results are that at the beginning, the model will perform poorly because it has not seen the whole dataset yet, and as more sessions are added to the dataset, the model will perform better.
    \item This process is repeated 10 times to have a 10-fold validation statistics.
\end{enumerate}

\subsection{Results}
An individual AU-NN is trained for each of the 27 subjects. Several hyperparameters were tested, and the one that performed the best was:
\begin{itemize}
    \item Architecture: Two layers, the first one (hidden) with 10 neurons, and the second one (last) with 2 neurons (1 per class), The number of membership functions for all the ANFIS-Units (neurons) is fixed to 3.
    \item The learning rate is fixed to 0.001.
    \item The epochs per session is fixed to 300
    \item The triangular membership function is used.
    \item The Adam optimizer is used to train the network.
\end{itemize}

As we are dealing with an IL problem, we need to study several metrics in each session. Also, because trials are not symmetric, it means that the test set will be unbalanced. To face this problem, the metrics chosen to be studied are: Train accuracy, Test accuracy, F1 score, Sensitivity, Specificity, and a metric defined as  (Sensitivity + Specificity)/2.

The sensitivity and specificity are included because the solely F1 score does not give the same weight to both classes. Thus, we know that the specificity focuses on true negatives, while the sensitivity focuses on true positives, giving with this a more deep understanding of how the model is performing.

Also, another independent AU-NN per subject is included, this other AU-NN is trained with all the dataset at once, this is done to have a reference of what is the maximum value we can expect the model to learn with the whole dataset. The hyperparameters for this second AU-NN are the same as the first AU-NN, but only the epochs are changed, since there are no sessions for this second AU-NN, the epochs are fixed to 1200.

Similarly, a Random Forest (RF) trained with all the dataset is also included, this is done to have a reference of a traditional approach of simpler machine learning models without IL option. The RF is chosen because in \cite{Hernandez-Del-Toro2021} is the one that reported better performance with this dataset and feature set.

\begin{figure}[ht]

\begin{minipage}{.33\linewidth}
\centering
\subfloat[]{\label{main:a}\includegraphics[scale=.27]{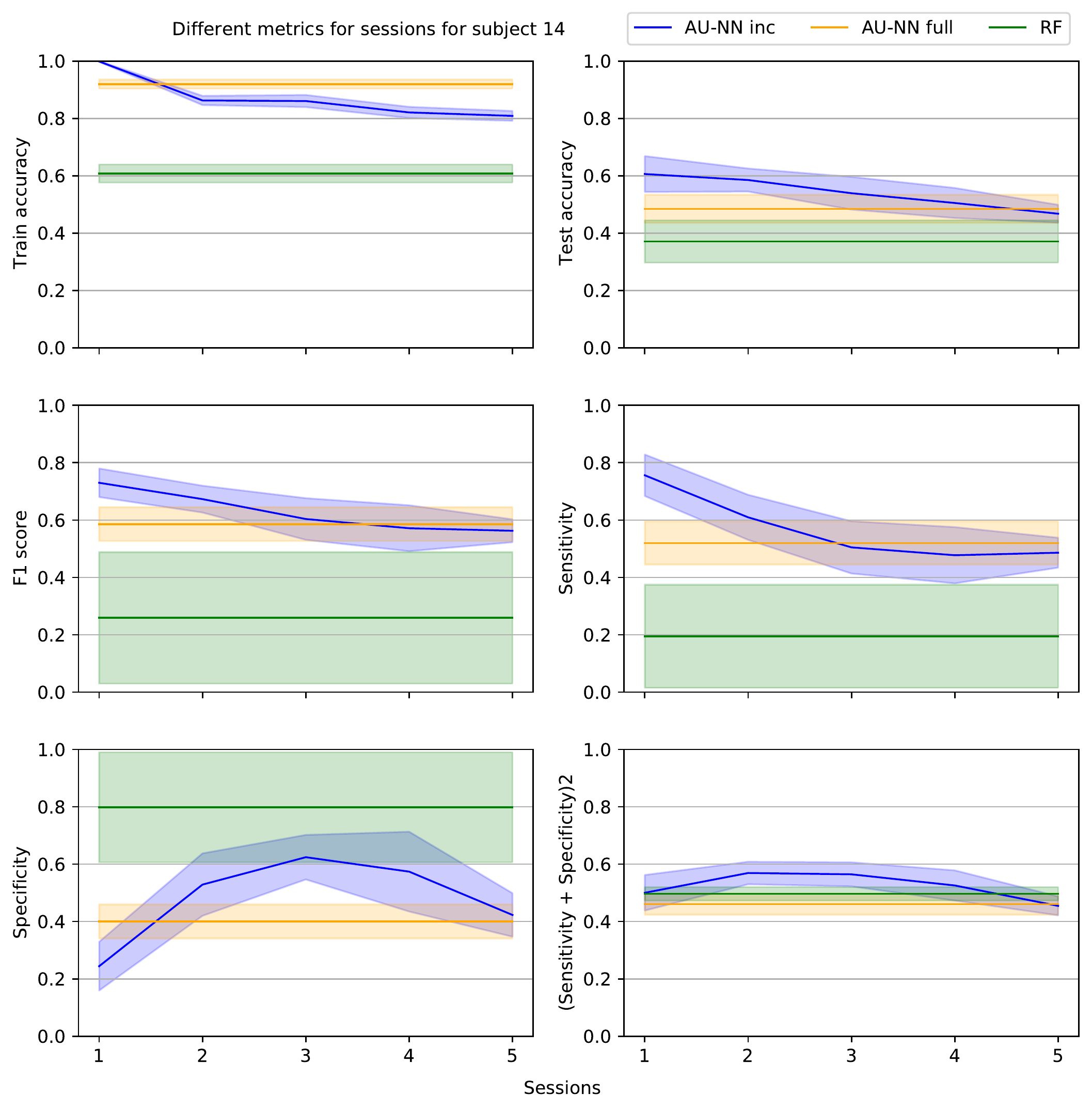}}
\end{minipage}%
\begin{minipage}{.33\linewidth}
\centering
\subfloat[]{\label{main:b}\includegraphics[scale=.27]{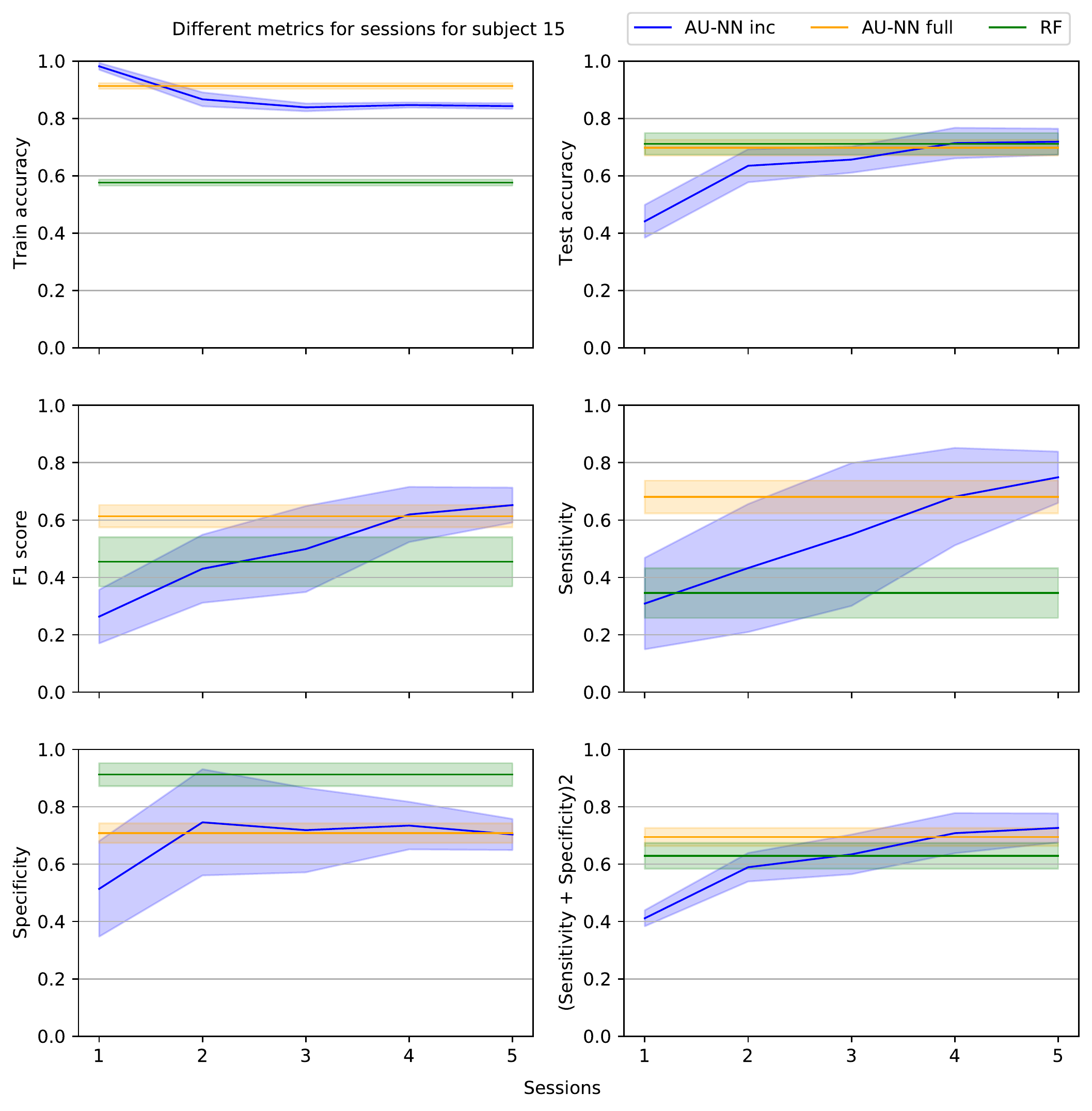}}
\end{minipage}%
\begin{minipage}{.33\linewidth}
\centering
\subfloat[]{\label{main:c}\includegraphics[scale=.27]{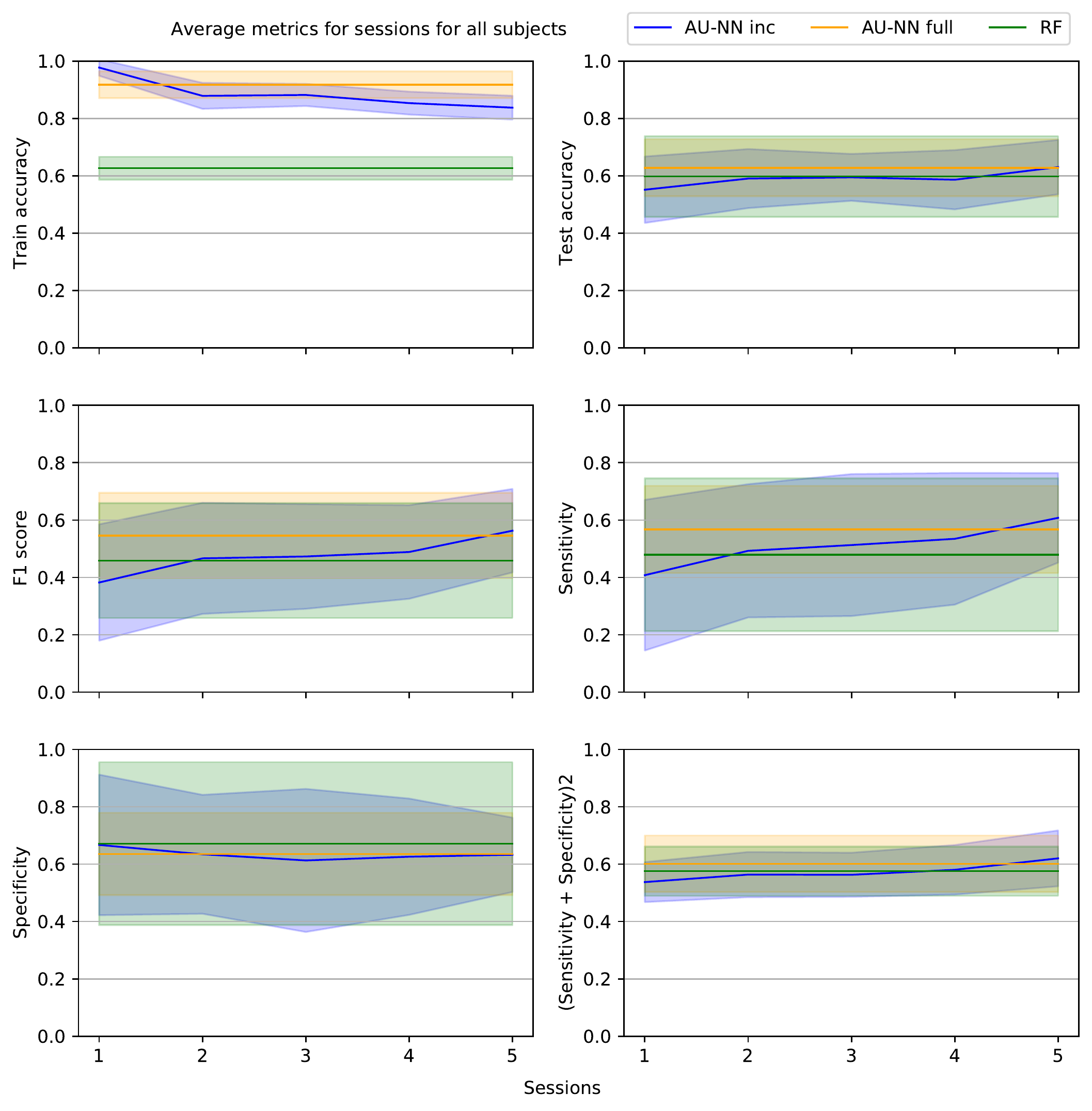}}
\end{minipage}%

\caption{Results obtained for the 3 more relevant subjects. (a) Subject 14: The worst performing subject. (b) Subject 15: The best performing subject. (c) Subject 28: The average over all subjects.}
\label{fig:IL_results}
\end{figure}

With the mentioned above, there are 6 distinct plots per subject, each one showing the metrics evolving with each session. In \autoref{fig:IL_results} are shown the results for three subjects which we selected as the most representative over all subjects:
\begin{itemize}
    \item Subject 14: The worst performing subject.
    \item Subject 15: The best performing subject.
    \item Subject 28: The average over all subjects.
\end{itemize}
The terms  ``worst performing" and ``best performing" are defined as the ones that behave as we expected (to perform poorly at the beginning and increase the metric in each session). The results of the rest of the subjects can be found at the github repository \url{https://github.com/tonahdztoro/AU_NN/tree/main/incremental_IWS_vs_ISS/Plots}.

\newpage
\subsection{Discussion}
In each of the 6 plots per subject, the x-axis represents the sessions (5), while the y-axis represents the obtained value for each metric. Also there are three colors, the blue color represents the performance of the AU-NN in an IL fashion. The orange color represents the AU-NN trained with the whole dataset at once, and the green color represents the RF trained with the whole dataset at once. In all three colors, the solid line represents the mean value of the 10-fold validation scheme, while the shaded area represents the standard deviation over the folds.

We can observe that in figure \autoref{main:a}, the metrics decrease over each session. However, the metrics obtained by the IL AU-NN end up at the same or above the results obtained by the AU-NN, and the RF trained with the whole dataset. For figure \autoref{main:b} we can see that with each session, the metrics increase, and in some cases end up above the result obtained by the AU-NN, and the RF trained with the whole dataset. In figure \autoref{main:c}, the average of the subjects, at first glance the figure looks very noisy, but on a detailed observation, we can see that for all the metrics, the value always rises, or maintains in each session.

For the rest of the subjects. In some metrics the results behave as expected, while in others don't. This can be solved by tuning specific hyperparameters per subject. In \cite{Hernandez-Del-Toro2021} it is reported that each subject behaves better with different feature sets. This could suggest also the same behavior for the models. Finding a specific architecture per subject can lead to a better performance.

We can see that in all subjects shown, the train accuracy decreases over session, this is due to the increasing of the dataset in each session. At the beginning, the model learns to detect a variety of ISS, but in the IWS instances, there is only included one word, and with the time, more of the five words are included, gaining with this, robustness in the model.

With this results we prove the capability of the AU-NN to deal with complex IL problems.

\section{Conclusion}\label{sec:conclusion}
We presented the AU-NN: ANFIS Unit Neural Network, a neural network that has independent ANFIS for neurons. This network was tested in two scenarios related with chaotic, and non-linear signals as EEG, where fuzzy systems are proven to perform good. the two scenarios are: (i) Classification of 5 imagined words, and (ii) Classification of IWS vs. ISS in an IL way. In the first scenario, the AU-NN outperformed the previous reported results for the selected dataset. In the second scenario, as there is no previous work involving IWS vs ISS in an IL fashion, the AU-NN is compared against traditional approaches as the RF using 6 metrics, and against another AU-NN which is trained with all the dataset at once. The results show that the incremental AU-NN performs good on average for the subjects, and in each session maintains the metrics or increase them. It is important to notice that in both cases, the networks performed good using architectures of small neural networks (1 layer with 5 neurons, and 2 layers with 10 and 2 neurons), respectively for each problem. Reinforcing with this, the capability of the AU-NN to perform better than a traditional neural network.

This shows the capability of the AU-NN of both dealing with complex problems as those involving EEG signals, and the capability of performing good with a dataset that requires IL. The codes for the implementation of the AU-NN as well as the codes used in these experiments are uploaded on a github repository for future uses.

\begingroup
\bibliographystyle{unsrt}
\bibliography{bib}
\endgroup

\end{document}